\def\year{2020}\relax
\documentclass[letterpaper]{article} 
\usepackage{aaai20}  
\usepackage{times}  
\usepackage{helvet} 
\usepackage{courier}  
\usepackage[hyphens]{url}  
\usepackage{graphicx} 
\usepackage{graphicx}  
\usepackage{amsthm}

\usepackage{amsmath}
\DeclareMathOperator*{\argmax}{arg\,max}

\title{Solving QSAT problems with neural MCTS}
\author{Ruiyang Xu and Karl Lieberherr\\ 
Khoury College of Computer Sciences \\ Northeastern University\\ 
ruiyang@ccs.neu.edu\\lieber@ccs.neu.edu 
}

\begin{document}

\maketitle

\begin{abstract}
Recent achievements from AlphaZero using self-play has shown remarkable performance on several board games. It is plausible to think that self-play, starting from zero knowledge, can gradually approximate a winning strategy for certain two-player games after an amount of training. In this paper, we try to leverage the computational power of neural Monte Carlo Tree Search (neural MCTS), the core algorithm from AlphaZero, to solve Quantified Boolean Formula Satisfaction (QSAT) problems, which are PSPACE complete. Knowing that every QSAT problem is equivalent to a QSAT game, the game outcome can be used to derive the solutions of the original QSAT problems. We propose a way to encode Quantified Boolean Formulas (QBFs) as graphs and apply a graph neural network (GNN) to embed the QBFs into the neural MCTS. After training, an off-the-shelf QSAT solver is used to evaluate the performance of the algorithm. Our result shows that, for problems within a limited size, the algorithm learns to solve the problem correctly merely from self-play. 
\end{abstract}

\section{Introduction}\label{introduction}
The past several years have witnessed the progress and success of self-play. The combination of classical MCTS \cite{mcts_survey} algorithms with newly developed deep learning techniques gives a stunning performance on complex board games like Go and Chess \cite{silver2018general,alpha0,alpha_go}. One common but outstanding feature of such an algorithm is the tabula-rasa style of learning. In other words, it learns to play the game with zero knowledge (except the game rules). Such tabula-rasa learning is regarded as an approach to general artificial intelligence.

Given such an achievement, it is interesting to see whether their algorithm's superhuman capability can be used to solve problems in other domains. Specifically, we apply neural MCTS \cite{silver2018general,expertit} to solve the QSAT problem through self-play on the QSAT game. Our experiment shows that, even though the QSAT game is fundamentally different from traditional board games (see section \ref{arch}), the algorithm is still able to determine the truthfulness of the corresponding QSAT problem through the dominant player. Furthermore, the trained algorithm can be used to approximate the solution (or show the non-existence of a solution) of the original problem through competitions against an enumerator. However, our objective is not necessarily to improve the state-of-the-art of hand-crafted problem solvers in specific areas but to illustrate that there is a generic algorithm (neural MCTS) that can solve well-known problems tabula-rasa.

In this work, we make two main contributions: 1. We propose a way to turn QBFs into graphs so that they can be embedded with a graph neural network; 2. We implemented a variant of the neural MCTS algorithm, which has two independent neural networks (designed explicitly for the QSAT games) for each of the player. Our result shows that the algorithm can determine the truthfulness of the QSAT problem correctly. The remainder of this paper is organized as follows.
Section \ref{relatedworks} shows some related works which inspired our work. Section \ref{preliminaries} presents essential preliminaries on neural MCTS, the QSAT problems, and graph neural networks. Section \ref{implement} introduces our approach to encode QBFs as graphs and the architecture of our implementation.
Section \ref{experiment} gives our correctness measurement and presents experimental results. Section \ref{discuss} and \ref{conclusion} made a discussion and conclusions.

\section{Related Work}\label{relatedworks}
In terms of combining a QSAT solver with machine learning, Janota built a competitive QSAT solver, QFUN \cite{qfun}, based on counterexample guided refinement and machine learning. Although like in our work, the QSAT problems is treated as a game, their learning does not depend on the game state (i.e., the QBF), but focus on the assignment pairs from the two players in two consecutive moves (i.e., a move by the existential player, and a countermove by the universal player). By supervised learning a decision tree classifier, the learning algorithm categorizes the move-countermove pairs into two classes: feasible countermove and infeasible countermove. QFUN progressively solves a QBF by learning moves for the existential player so that there are no feasible countermoves for the universal player. While the performance is compelling, their solver is largely based on a counterexample guided abstraction refinement algorithm \cite{CEGAR}, whose design requires insights from human, hence cannot be regarded as tabula-rasa.  

As an alternative methodology, NeuroSAT \cite{selsam2018learning} provides us another insight to apply machine learning on such problems. By leveraging the graph neural networks \cite{battaglia2018relational} and message passing process \cite{gilmer2017neural}, they developed a single-bit supervised SAT solver. The algorithm depends on zero-knowledge and learns purely on the input formula. In NeuroSAT, Boolean formulas are encoded as graphs so that a specially designed graph neural network can be applied to those graphs. The target value of the graph neural network is a single bit, represented for the solvability of the input SAT problem. It has been shown that NeuroSAT performs adequately on SAT problems within a reasonable size. 

When it comes to applying neural MCTS to solve problems in other domains, Xu et al. use a technique called Zermelo Gamification to turn specific combinatorial optimization problems into games so that they can be solved through AlphaZero like algorithms \cite{ruiyang2019}. They applied their method to a particular combinatorial optimization problem called HSR. Their result shows that the algorithm can accurately solve such a problem within a given size. Although they only applied their method to one specific problem, their experiment result endorse the idea that there is a generic algorithm (neural MCTS) that can solve well-known problems tabula-rasa. To this extent, our work can be seen as an extension of theirs. 

\section{Preliminaries}\label{preliminaries}
\subsection{Neural Monte Carlo Tree Search}\label{mcts}
The PUCT (Predictor +
UCT) algorithm implemented in AlphaZero \cite{alpha0,gochessshogi} is essentially a neural MCTS algorithm which uses PUCB Predictor +
UCB \cite{Rosin2011} as its confidence upper bound \cite{uct} and uses the neural prediction $P_{\phi}(a|s)$ as the predictor. The algorithm is running through multiple searching iterations to decide the optimal action for the current state. During each iteration, there are 4 phases:
\begin{enumerate}
\item{SELECT: }
At the beginning of each iteration, the algorithm selects a path from the root (current game state) to a leaf (either a terminal state or an unvisited state) in the tree according to the PUCB (see \cite{alpha_go} for a detailed explanation for terms used in the formula). Specifically, suppose the root is $s_0$, we have \footnote{Theoretically, the exploratory term should be $\sqrt{\frac{\sum_{a'}N(s_{i-1},a')}{N(s_{i-1},a)+1}}$, however, the AlphaZero used the variant $\frac{\sqrt{\sum_{a'} N(s_{i-1},a')}}{N(s_{i-1},a)+1}$ without any explanation. We tried both in our implementation, and it turns out that the AlphaZero one performs much better.}:
$$a_{i}=\argmax_a\left[Q(s_{i},a)+cP_\phi(a|s_{i})\frac{\sqrt{\sum_{a'} N(s_{i},a')}}{N(s_{i},a)+1}\right]$$
$$Q(s_{i},a) = \frac{W(s_{i},a)}{N(s_{i},a)+1}$$
$$s_{i+1}=move(s_{i},a_{i})$$
\item{EXPAND: }
Once the select phase ends at a non-terminal leaf, the leaf will be fully expanded and marked as an internal node of the current tree. All its children nodes will be considered as leaf nodes during the next iteration of selection.
\item{ROLL-OUT: }
Normally, starting from the expanded leaf node chosen from previous phases, the MCTS algorithm uses a random policy to roll out the rest of the game \cite{mcts_survey}. The algorithm simulates the actions of each player randomly until it arrives at a terminal state, which means the game has ended. The algorithm then uses the outcome of the game as the result evaluation for the expanded leaf node.

However, a random roll-out usually becomes time-consuming when the tree is deep. A neural MCTS algorithm, instead, uses a neural network $V_{\phi}$ to predict the result evaluation so that the algorithm saves the time on rolling out.   

\item{BACKUP: }
This is the last phase of an iteration where the algorithm recursively backs-up the result evaluation in the tree edges. Specifically, suppose the path found in the Select phase is $\{(s_0,a_0),(s_1,a_1),...(s_{l-1},a_{l-1}),(s_l,\_)\}$. then for each edge $(s_i,a_i)$ in the path, we update the statistics as:
$$W^{new}(s_i,a_i)=W^{old}(s_i,a_i)+V_{\phi}(s_l)$$
$$N^{new}(s_i,a_i)=N^{old}(s_i,a_i)+1$$
However, in practice, considering a Laplace smoothing in the expression of Q, the following updates are actually applied:
$$Q^{new}(s_i,a_i)=\frac{Q^{old}(s_i,a_i)\times N^{old}(s_i,a_i)+V_{\phi}(s_l)}{N^{old}(s_i,a_i)+1}$$
$$N^{new}(s_i,a_i)=N^{old}(s_i,a_i)+1$$

Once the given number of iterations has been reached, the algorithm returns a vector of action probabilities of the current state (root $s_0$). And each action probability is computed as $\pi(a|s_0)=\frac{N(s_0,a)}{\sum_{a'}N(s_0,a')}$. The real action played by the neural MCTS is then sampled from the action probability vector $\pi$. In this way, neural MCTS simulates the action for each player alternately until the game ends. This process is called neural MCTS simulation, which is the core of self-play. 
\end{enumerate}

\subsection{QSAT Problems and QSAT games}\label{qsat}
A quantified Boolean formula (QBF) is a formula in the following form:
$$\exists x_1 \forall x_2...\exists x_n.\Phi(x_1,...,x_n)$$
Where $x_i$ are distinct boolean variables. The sequence of quantifiers and variables is called the prefix of a QBF.  The propositional formula $\Phi$ is called the matrix of a QBF, which only uses the variables in $\{x_i\}$. A QBF can evaluate to either true or false since there are no free variables, and it is solvable only if it evaluates to true, otherwise, it is unsolvable. The problem of determining the truthfulness of a QBF is called QSAT problem, which is known to be PSPACE complete.  

A QSAT problem can be seen as a game between two players: the existential player (the Proponent (P)) who assigns values to the existentially quantified variables, and the universal player (the Opponent (OP)) who assigns values to the universally quantified variables. The two players make moves by assigning values to the variables alternately following the sequence of quantifiers in the prefix. The existential player (P) wins if the formula evaluates to True and the universal player (OP) wins if it evaluates to False.

\subsection{Gated Graph Neural Networks}\label{gnn}
In this work, QBFs are encoded as graphs, and a Gated Graph Neural Network (GGNN) \cite{ggnn,gilmer2017neural} is applied to embed the QBFs into the neural MCTS framework. Notice that the GGNN is not the only option and there are alternatives \cite{gilmer2017neural,battaglia2018relational}, we choose GGNN for the sake of its easy implementation.  

The forward pass of the GGNN can be described as following:
$$m_v^{t+1}=\sum_e\sum_{w\in N(v)}A_{e_{wv}}h_w^t,t=0..T$$
$$h_v^{t+1}=GRU(h_v^t,m_v^{t+1}),t=0..T$$
$$R=\sum_{v\in V}\sigma(f(h_v^T,h_v^0))\odot g(h_v^T)$$
where $e$ is the edge type in a multigraph, $A_e$ is the edge-weight matrix to be learned during the training, $h_v^t$ is the hidden representation of node $v$ at message passing iteration $t$, and $m_v^t$ is called the message from node $v$ at iteration $t$. $R$ is called the read-out which aggregates information from each node to generate a global feature target (notice that $\sigma$ means the sigmoid activation function, $f$ and $g$ are MLPs, and $\odot$ means element-wise product). 

The message passing process iterates for a given $T$ times, during each iteration, each node $v$ computes its message using the hidden representation from the neighbor nodes $N(v)$. After that, a Gated Recurrent Unit (GRU) is used to update the hidden representation of the node $v$. The message passing process allows each node's hidden representation to capture the global structure information of the entire input graph. Finally, the read-out process $R$ is applied to all the nodes to compute the global target of the input graph. GGNN is invariant to graph isomorphism, which is well-suited to capture the symmetry properties among the QBFs.

\section{Implementation}\label{implement}
\subsection{QBF Graphs}
Although the QSAT problem has a simple syntactic structure, symmetries induced by the semantics of propositional logic should not be ignored \cite{qbfsym}. The fact that symmetric QBFs are equivalent can improve learning efficiency. In this work, we specially designed a graph encoding of the QBFs, which helps us catch those symmetries through graph isomorphism. 

\begin{figure}[ht]
\centering
\includegraphics[width=0.4\textwidth]{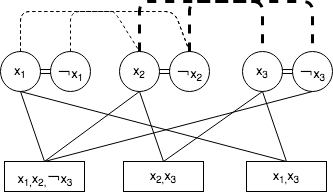}
\caption{An example of graph encoding for the QBF: $\exists x_1\forall x_2\exists x_3 (x_1\vee x_2\vee \neg x_3)\wedge(x_2\vee x_3)\wedge(x_1\vee x_3)$. Notice that there are four types of edges, and two types of nodes.}
\label{qbfg}
\end{figure}

After using Tseyting transformation to re-write $\Phi$ in conjunctive normal form (CNF), a QBF is represented as an undirected multigraph (Fig. \ref{qbfg}) with two nodes for every literal and its negation, and one node for every clause. There are four types of edges in this multigraph: 1. E2A edge, an edge between every consecutive existential literal and universal literal; 2. A2E edge, an edge between every consecutive universal literal and existential literal; 3. L2C edge, an edge between every literal and every clause it appears in; 4. reflexive edge, and an edge between each pair of literal and its negation.

The reason behind such a design are three aspects: 1. the sequential information of the prefix is essential to identify the solution of a QBF. Even if two QBFs have the same matrix $\Phi$, a different variable sequence in the prefix might lead to a massive difference in the solution. Therefore, we use the E2A edges and A2E edges to track such sequential information. 2. In a QBF, variables only show as positive literals in the prefix; however, they can be both positive and negative in the matrix $\Phi$. Hence we naturally represent any variable as two nodes, meaning a pair of two complementary literals. 3. Since any literal and its complement are coupled, we use a reflexive edge to capture such entanglement. 

\subsection{Architecture}\label{arch}
In our design, the policy-evaluation neural network of the neural MCTS becomes two GGNNs (see section \ref{gnn}), one for each player. The reason why we use two independent neural networks instead of one is that the QSAT game is asymmetric in terms of the winning condition. As we have introduced in the section \ref{qsat}, P wins the game if and only if the QBF evaluates to true, while OP wins the game if and only if the QBF evaluates to false. On the other hand, when it comes to the outcome of GGNN for two consecutive moves by different players, we noticed that the hidden representation sometimes has no significant difference between the two players. Hence the GGNN becomes confused on the input graphs. This issue can be resolved only by separating the neural networks, so that both of the players can learn and progress mutually and consistently.

Another fact to notice is that we treat every single QSAT problem as an independent game. During the self-play phase, the neural MCTS algorithm (section \ref{mcts}) simulates the move for each player based on the player's GGNN. The neural MCTS takes in the current game state (the QBF graph) and uses the current player's GGNN to do the selection and rollout. After a certain number (25 in our case) of iterations, neural MCTS will return the action probability distribution for the current state. The player will sample her next move from this distribution. The simulation alternates between the two players until the game ends, where the game outcome will be evaluated and stored for the training phase. 
To call the neural network, the hidden representation $h_v^0$ of each node $v$ is initialized with the type of the node. Specifically, for an existential literal node, the hidden representation is $[1,0,0,...,0]$; for a universal literal node, the hidden representation is $[0,1,0,...,0]$; and for a CNF clause node, the hidden representation is $[0,0,1,...,0]$ . Notice that we use $0$'s to pad the vector to a given length. Another fact to notice is that  there are two read-out task ($P_{\phi}$ and $V_{\phi}$). Hence we use two different sets of aggregation MLPs for each of the task:
$$R_i=\sum_{v\in V}\sigma(f_i(h_v^T,h_v^0))\odot g_i(h_v^T)$$
$$P_{\phi}=R_1,V_{\phi}=R_2$$

After each self-play simulation, we store the game trace of each player separately as a set of tuples in the form of ($s$, $\pi$, $v$), where $s$ is the game state (the QBF graph), $\pi$ is the action probability distribution generated by neural MCTS based on current state, and $v$ is the game result in the perspective of current player based on the game outcome. We run such a simulation several times (in our case, ten times) to retrieve enough training data. After that, we train the GGNN independently for each of the players using those training data collected during self-play. After training, we use the newly trained GGNNs to play against each other for 20 rounds and collect the performance data for evaluation and analysis, and this is called the arena phase. 

\section{Experiment}\label{experiment}
\subsection{Experiment Setup}
The hyperparameters are set as follows:  the number of searching iteration for neural MCTS is set to 25, and the number of simulation is set to 100; the message passing time $T$ is set to 10 for the GGNN; the size of the hidden representation of the GGNN is set to 128.

Considering the capacity and computation power of our machine, we generate 20 random QBFs (10 solvable and 10 unsolvable) which have 51 nodes (the prefix has 21 quantified variables, and the matrix has 9 clauses. So there are 42 literal nodes and 9 clause nodes.) after encoding as graphs.  
Each QBF is regarded as a single game to be played and learned by the neural MCTS. We run the learning iteration (i.e., self-play, training, and arena) for 32 epochs, and collect the performance data in the arena phase during each iteration. 

\subsection{Performance Measurement}
To measure the performance of the algorithm, we use two metrics: the local correctness ratio and the global correctness ratio. We compute the local correctness ratio of the two players during the arena phase where the two players compete with each other for 20 rounds. The action is locally correct if it preserves a winning position. It is straightforward to check the local correctness of actions using a QSAT solver: GhostQ \cite{ghostq}. We collect the local correctness ratio of the two players after each round of competing in arena phase. Then we take the average value of their local correctness ratio as the performance measurement for the current training iteration.

\theoremstyle{definition}
\newtheorem{definition}{Definition}[section]

\theoremstyle{definition}
\begin{definition}{Local Correctness for P}
\\Given a QBF $\exists x_1 \forall x_2...\exists x_n.\Phi$, an action $x^*$ is locally correct if and only if $\forall x_2...\exists x_n.\Phi[x_1\setminus x^*]$ evaluates to True.
\end{definition}
\begin{definition}{Local Correctness for OP}
\\Given a QBF $\forall x_1 \exists x_2...\exists x_n.\Phi$, an action $x^*$ is locally correct if and only if $\exists x_2...\exists x_n.\Phi[x_1\setminus x^*]$ evaluates to False.
\end{definition}

Since the two neural networks might be inductively biased to each other, the locally correct solution could be incorrect. To see whether the neural MCTS learns the correct solution, we measure the global correctness ratio by test the algorithm with an enumerator. To be specific, if a QBF is satisfactory, then we enumerate all possible move for the OP (the universal player) and use the enumerator to play against the P's neural network. Vice-versa for the unsatisfactory QBF. Theoretically, OP's neural network fails to solve the QBF if there is any chance that the P's enumerator can win the game. We count the number of winning games for each player and use it to measure the global correctness ratio. A 100\% global correctness not only means the neural MCTS has found the correct solution, but also a fully support (represented as a winning strategy encoded in the neural network) to that solution. On the other hand, a non-100\% global correctness can be treated as a measure of approximation of the algorithm.

\subsection{General Result}
Our experiment shows that the algorithm can correctly determine the truthfulness of all 20 test QBFs. We notice that, for a solvable QBF, the existential player can quickly dominate the game and win at most of the times, and vice-versa for the universal player in an unsolvable case. The result indicates that for a solvable/ an unsolvable QSAT problem, the existential/ universal player has a higher chance to win the corresponding QSAT game against the universal/ existential player.

We also measured the algorithm's global correctness ratio for all test cases, and we noticed an asymmetry between the solvable and unsolvable cases. To be specific, we computed the average global correctness ratio (AGC) for all solvable and unsolvable QBFs respectively, and it turns out that the AGC for solvable cases is 87\%, while the AGC for unsolvable cases is 85\%. This fact indicates that neural MCTS can still be an adequate approximator to QSAT problem, even if it cannot derive a 100\% correct strategy. 

\subsection{Two Examples}\label{examples}
\begin{figure}[ht]
\centering
\includegraphics[width=0.4\textwidth]{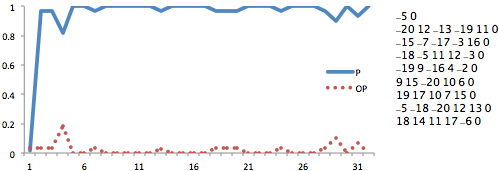}
\caption{Local correctness ratio measured for a solvable QBF. The matrix of the QBF is listed on the right side in QDIMACS format.}
\label{qbfsat}
\end{figure}

\begin{figure}[ht]
\centering
\includegraphics[width=0.4\textwidth]{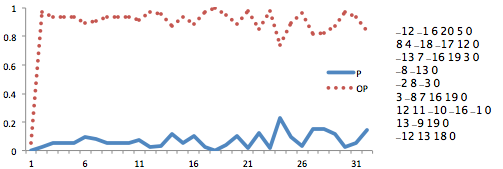}
\caption{Local correctness ratio measured for an unsolvable QBF. The matrix of the QBF is listed on the right side in QDIMACS format.}
\label{qbfunsat}
\end{figure}

In this section, for the illustration purpose, we show the experiment results for a solvable QSAT and an unsolvable QSAT (described in Fig. \ref{qbfsat} and Fig. \ref{qbfunsat} where, due to limited space, we only show the matrix of the QBF). One can see, in Fig. \ref{qbfsat}, the local correctness ratio of the existential player (P) soars high after the first epoch; while in Fig. \ref{qbfunsat}, the local correctness ratio of the universal player (OP) increases rapidly. Even though there are fluctuations, one of the player always dominate the game, this phenomenon is treated as an indicator to the truthfulness of the QSAT. Also, notice that the curves in the unsolvable case wave more violently than the ones in the solvable case. This fact means that even though the player can dominate the game, dominating an unsolvable QSAT game might be harder than a solvable one. In terms of global correctness ratio, both of them got 100\% correctness, that means the neural MCTS not only makes the correct decision but also constructively support its decision. 


\section{Discussion}\label{discuss}

\subsection{Exploration v.s. Exploitation} 
One of the known issues of self-play is that the two players will always mutually bias their strategy to fit with the other's one through exploiting their experiences. This mutual inductive bias facilitates the learning process of the players when they are at the same pace. However, once the learning speeds are unbalanced, the mutual inductive bias foils the improvement of players' performance by stagnating their strategies in a local optimum. To understand this issue, one can think about a game between an expert and a newbie. Since the expert can easily find a strategy to win against the newbie, the newbie will always lose the game. And because there is no positive feedback at all, the newbie will build a biased belief that there is no way to win the game. Such a belief can be strengthened during self-play, and finally, it leads to some fixed losing strategy. While on the other side, since the opponent is not so challenging, the expert will also stay with the current strategy without any attempt to improve it.

Nevertheless, we notice that neural MCTS is resilient to mutual inductive bias. Whenever the learning paces are unbalanced, the weaker players decisions become indifferent (i.e., no matter what moves she takes, she will always lose the game). On the other hand, neural MCTS pushes those indifferent actions into a uniform distribution, hence to encourage exploration by making random moves. Consequently, neural MCTS adaptively balance the exploration and exploitation, thus jumping out of the local optimal.     

\subsection{State Space Coverage}
\begin{figure}[ht]
\centering
\includegraphics[width=0.4\textwidth]{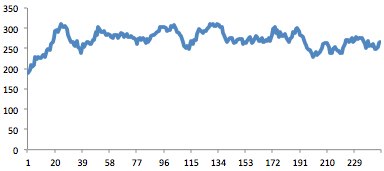}
\caption{Average states accessed during self-play for QSAT problem described in Fig. \ref{qbfsat}. As a comparison, there are 226599 states in total.}
\label{coverage}
\end{figure}

Neural MCTS is capable of handling a large state space \cite{alpha0}. Such an algorithm must search only a small portion of the state space and make the decisions from those limited observations. To measure the state space coverage, we recorded the number of states accessed during the experiment, In each QSAT game, we count the total number of states accessed during each game in self-play, and we compute the 10-game moving average of state accessed for all self-plays (Fig. \ref{coverage}). This result indicates an implicit adaptive pruning mechanism behind the neural MCTS algorithm, which can be regarded as a justification for its capability of handling large state spaces.

\subsection{Limitation}
Our test cases are restricted to a limited size. Because QSAT is known to be PSPACE complete, verifying the correctness of the algorithm is time-consuming. In our experiment, there are 10 to 11 moves for each player. Hence to verify the correctness of the algorithm, it roughly takes $2^{10}$ to $2^{11}$ tests. And the verification time increases exponentially with the number of variables in the QBF.

On the other hand, the strategy learned by the neural MCTS algorithm is implicitly encoded inside the neural network, and there is no way to extract such a strategy so that they can be explicitly verified by any more efficient approaches from the formal method. Therefore, using an enumerator to verify the correctness is inevitable for the time being. As a result, even though neural MCTS can handle a deep game tree hence a large number of variables, it is still hard or even impossible to verify the learning outcome.  

\section{Conclusion}\label{conclusion}
In this work, intrigued by the astonishing achievements from AlphaZero, we attempt to leverage the computational power of neural MCTS algorithm to solve a practical problem: QSAT. We make two main contributions. First, we propose a way to encode QBFs as undirected multigraphs, which bridges the logic formula representation of QBFs with the graph neural network input. Second, we particularly use two separated graph neural networks to build our neural MCTS variant. Such a design can significantly reduce the learning confusion caused by the asymmetry between the two players. Our evaluation is based on two metrics: local and global correctness ratio. The local metric, by utilizing an off-the-shelf QSAT solver, only focus on the correctness in a single game, yet it imposes no constraints on the number of variables in the formula; The global metric, relies on an enumerator, can determine the exact correctness of the learned neural MCTS, but it is sensitive to the number of variables in the formula. Our experiment results are positive on the given limited size test cases, which justifies the feasibility of our idea to some extents. For future work, it may be worthwhile to figure out how to explain the learned neural MCTS or how to extract the generated strategy from the neural network. It is also useful to do some study on how to optimize the current algorithm so that it can handle more significant cases. Our objective is not necessarily to improve the state-of-the-art of hand-crafted problem solvers in specific areas but to illustrate that there is a generic algorithm (neural MCTS) that can solve well-known problems tabula-rasa. The hope is that neural MCTS will help solve future algorithmic problems that have not yet been solved by humans. We view Neural MCTS as a helper in human solving of algorithmic problems in the future. We also hope our research sheds some light on the remarkable but mysterious learning ability of the neural MCTS algorithm from AlphaZero.

\bibliographystyle{aaai}
\bibliography{references}

\begin{thebibliography}{}

\bibitem[\protect\citeauthoryear{Anthony, Tian, and Barber}{2017}]{expertit}
Anthony, T.; Tian, Z.; and Barber, D.
\newblock 2017.
\newblock Thinking fast and slow with deep learning and tree search.
\newblock In {\em Advances in Neural Information Processing Systems},
  5360--5370.

\bibitem[\protect\citeauthoryear{Battaglia \bgroup et al\mbox.\egroup
  }{2018}]{battaglia2018relational}
Battaglia, P.~W.; Hamrick, J.~B.; Bapst, V.; Sanchez-Gonzalez, A.; Zambaldi,
  V.; Malinowski, M.; Tacchetti, A.; Raposo, D.; Santoro, A.; Faulkner, R.;
  et~al.
\newblock 2018.
\newblock Relational inductive biases, deep learning, and graph networks.
\newblock {\em arXiv preprint arXiv:1806.01261}.

\bibitem[\protect\citeauthoryear{Browne \bgroup et al\mbox.\egroup
  }{2012}]{mcts_survey}
Browne, C.; Powley, E.~J.; Whitehouse, D.; Lucas, S.~M.; Cowling, P.~I.;
  Rohlfshagen, P.; Tavener, S.; Liebana, D.~P.; Samothrakis, S.; and Colton, S.
\newblock 2012.
\newblock {A} {Survey} of {Monte} {Carlo} {Tree} {Search} {Methods}.
\newblock {\em IEEE Trans. Comput. Intellig. and AI in Games} 4(1):1--43.

\bibitem[\protect\citeauthoryear{Gilmer \bgroup et al\mbox.\egroup
  }{2017}]{gilmer2017neural}
Gilmer, J.; Schoenholz, S.~S.; Riley, P.~F.; Vinyals, O.; and Dahl, G.~E.
\newblock 2017.
\newblock Neural message passing for quantum chemistry.
\newblock In {\em Proceedings of the 34th International Conference on Machine
  Learning-Volume 70},  1263--1272.
\newblock JMLR. org.

\bibitem[\protect\citeauthoryear{Janota \bgroup et al\mbox.\egroup
  }{2012}]{CEGAR}
Janota, M.; Klieber, W.; Marques-Silva, J.; and Clarke, E.
\newblock 2012.
\newblock Solving {QBF} with counterexample guided refinement.
\newblock In {\em Proceedings of the 15th International Conference on Theory
  and Applications of Satisfiability Testing}, SAT'12,  114--128.
\newblock Springer-Verlag.

\bibitem[\protect\citeauthoryear{Janota}{2018}]{qfun}
Janota, M.
\newblock 2018.
\newblock Towards generalization in {QBF} solving via machine learning.
\newblock In {\em Proceedings of the Thirty-Second {AAAI} Conference on
  Artificial Intelligence, (AAAI-18)},  6607--6614.

\bibitem[\protect\citeauthoryear{Kauers and Seidl}{2018}]{qbfsym}
Kauers, M., and Seidl, M.
\newblock 2018.
\newblock Symmetries of quantified boolean formulas.
\newblock {\em ArXiv} abs/1802.03993.

\bibitem[\protect\citeauthoryear{Klieber \bgroup et al\mbox.\egroup
  }{2010}]{ghostq}
Klieber, W.; Sapra, S.; Gao, S.; and Clarke, E.
\newblock 2010.
\newblock A non-prenex, non-clausal {QBF} solver with game-state learning.
\newblock In {\em Proceedings of the 13th International Conference on Theory
  and Applications of Satisfiability Testing}, SAT'10,  128--142.
\newblock Springer-Verlag.

\bibitem[\protect\citeauthoryear{Kocsis and Szepesvari}{2006}]{uct}
Kocsis, L., and Szepesvari, C.
\newblock 2006.
\newblock {Bandit} {Based} {Monte-Carlo} {Planning}.
\newblock In {\em ECML}, volume 4212 of {\em Lecture Notes in Computer
  Science},  282--293.
\newblock Springer.

\bibitem[\protect\citeauthoryear{Li \bgroup et al\mbox.\egroup }{2016}]{ggnn}
Li, Y.; Tarlow, D.; Brockschmidt, M.; and Zemel, R.~S.
\newblock 2016.
\newblock Gated graph sequence neural networks.
\newblock In {\em 4th International Conference on Learning Representations,
  {ICLR} 2016, San Juan, Puerto Rico, May 2-4, 2016, Conference Track
  Proceedings}.

\bibitem[\protect\citeauthoryear{Rosin}{2011}]{Rosin2011}
Rosin, C.~D.
\newblock 2011.
\newblock Multi-armed bandits with episode context.
\newblock {\em Annals of Mathematics and Artificial Intelligence}
  61(3):203--230.

\bibitem[\protect\citeauthoryear{Selsam \bgroup et al\mbox.\egroup
  }{2018}]{selsam2018learning}
Selsam, D.; Lamm, M.; Bunz, B.; Liang, P.; de~Moura, L.; and Dill, D.~L.
\newblock 2018.
\newblock Learning a sat solver from single-bit supervision.
\newblock {\em arXiv preprint arXiv:1802.03685}.

\bibitem[\protect\citeauthoryear{Silver \bgroup et al\mbox.\egroup
  }{2016}]{alpha_go}
Silver, D.; Huang, A.; Maddison, C.~J.; Guez, A.; Sifre, L.; van~den Driessche,
  G.; Schrittwieser, J.; Antonoglou, I.; Panneershelvam, V.; Lanctot, M.;
  Dieleman, S.; Grewe, D.; Nham, J.; Kalchbrenner, N.; Sutskever, I.;
  Lillicrap, T.; Leach, M.; Kavukcuoglu, K.; Graepel, T.; and Hassabis, D.
\newblock 2016.
\newblock {Mastering} the game of {Go} with deep neural networks and tree
  search.
\newblock {\em Nature} 529:484.

\bibitem[\protect\citeauthoryear{Silver \bgroup et al\mbox.\egroup
  }{2017a}]{gochessshogi}
Silver, D.; Hubert, T.; Schrittwieser, J.; Antonoglou, I.; Lai, M.; Guez, A.;
  Lanctot, M.; Sifre, L.; Kumaran, D.; Graepel, T.; Lillicrap, T.~P.; Simonyan,
  K.; and Hassabis, D.
\newblock 2017a.
\newblock {Mastering} {Chess} and {Shogi} by {Self-Play} with a {General}
  {Reinforcement} {Learning} {Algorithm}.
\newblock {\em CoRR} abs/1712.01815.

\bibitem[\protect\citeauthoryear{Silver \bgroup et al\mbox.\egroup
  }{2017b}]{alpha0}
Silver, D.; Schrittwieser, J.; Simonyan, K.; Antonoglou, I.; Huang, A.; Guez,
  A.; Hubert, T.; Baker, L.; Lai, M.; Bolton, A.; Chen, Y.; Lillicrap, T.; Hui,
  F.; Sifre, L.; van~den Driessche, G.; Graepel, T.; and Hassabis, D.
\newblock 2017b.
\newblock {Mastering} the game of {Go} without human knowledge.
\newblock {\em Nature} 550:354.

\bibitem[\protect\citeauthoryear{Silver \bgroup et al\mbox.\egroup
  }{2018}]{silver2018general}
Silver, D.; Hubert, T.; Schrittwieser, J.; Antonoglou, I.; Lai, M.; Guez, A.;
  Lanctot, M.; Sifre, L.; Kumaran, D.; Graepel, T.; et~al.
\newblock 2018.
\newblock A general reinforcement learning algorithm that masters chess, shogi,
  and go through self-play.
\newblock {\em Science} 362(6419):1140--1144.

\bibitem[\protect\citeauthoryear{Xu and Lieberherr}{2019}]{ruiyang2019}
Xu, R., and Lieberherr, K.
\newblock 2019.
\newblock Learning self-game-play agents for combinatorial optimization
  problems.
\newblock In {\em Proceedings of the 18th International Conference on
  Autonomous Agents and MultiAgent Systems}, AAMAS '19,  2276--2278.
\newblock IFAAMS.

\end{thebibliography}

\end{document}